\documentclass[journal,letterpaper]{IEEEtran}

\usepackage{amsmath,amssymb,amsfonts}
\usepackage{bm}
\usepackage{array}
\usepackage{textcomp}
\usepackage{stfloats}
\usepackage{url}
\usepackage{graphicx}
\usepackage{cite}
\usepackage[hidelinks]{hyperref}
\hypersetup{
  pdftitle={Pre-Execution Robot Energy Prediction in Milling via a Jacobian-Aware Current-First Cascade},
  pdfauthor={Yecheng Jiao, Chenglin Li, Kai Ren, Yunfeng Zhang},
  pdfsubject={Preprint submitted to IEEE Transactions on Automation Science and Engineering}}
\graphicspath{{Figures/}}

\newenvironment{IEEEpractitioners}
  {\renewcommand{\abstractname}{Note to Practitioners}\begin{abstract}}
  {\end{abstract}}

\begin{document}

\title{Pre-Execution Robot Energy Prediction in Milling via a
Jacobian-Aware Current-First Cascade}

\author{Yecheng Jiao, Chenglin Li, Kai Ren, and Yunfeng Zhang%
\thanks{This work has been submitted to the IEEE for possible publication. Copyright may be transferred without notice, after which this version may no longer be accessible.}%
\thanks{\textit{(Corresponding authors: Chenglin Li; Kai Ren.)}}%
\thanks{Yecheng Jiao, Chenglin Li, and Yunfeng Zhang are with the Department of Mechanical Engineering, National University of Singapore, Singapore, and also with the National University of Singapore (Chongqing) Research Institute, China (e-mail: lichenglin@u.nus.edu).}%
\thanks{Kai Ren is with the State Key Laboratory of Fluid Power and Mechatronic Systems, School of Mechanical Engineering, Zhejiang University, China, and also with the Zhejiang Key Laboratory of Additive Manufacturing Technology and Equipment, School of Mechanical Engineering, Zhejiang University, China (e-mail: ren\_kai@zju.edu.cn).}%
}

\markboth{}{}

\maketitle

\begin{abstract}
Even for the same milling task, robot energy consumption (EC) can vary across configurations, making pre-execution prediction important for energy-aware trajectory planning. However, time-varying cutting loads and their configuration-dependent transmission to the joints make accurate prediction challenging.
This paper presents a Jacobian-aware, current-first cascade for pre-execution robot power and energy prediction in milling. The framework combines a rigid-body dynamics-based current estimate, cutting-induced current (CIC) prediction using machining parameters and the transposed Jacobian, residual-current compensation, and power estimation. A simulation-kinematics and real-electrical (SK-RE) dataset pairs simulated joint kinematics with measured joint currents and robot power, enabling prediction from planned inputs without measured force or electrical signals during inference.
The model was trained and tested using physical robotic milling trials covering different machining features and cutting conditions. 
On the two unseen machining features, the power root mean squared
error (RMSE) averaged 0.781~W, 52.8\% lower than that of the best-performing standalone data-driven baseline. The mean per-trial energy errors were 0.88\% and 1.45\%. These results demonstrate the potential of pre-execution energy evaluation of robotic milling trajectories.
\end{abstract}

\begin{IEEEpractitioners}
In robotic milling, different robot configurations can follow the same toolpath yet consume different amounts of energy. Physically testing every candidate trajectory is impractical during offline planning. This paper provides an energy evaluator that accounts for changing cutting loads and their configuration-dependent effects on joint currents. By pairing simulated motion with measured electrical data during training, it predicts robot power and energy using only simulated motion and planned cutting parameters. Following initial dynamics identification, the model was trained using only about 2.2 hours of cutting data from representative one-way slot and U-slot trials covering different robot postures and cutting conditions. Tests on representative unseen combinations of machining geometries and operating conditions achieved a mean power RMSE 52.8\% lower than that of the best-performing standalone data-driven baseline.

The underlying dynamics and load-transmission relations are formulated for general serial manipulators, with robot-specific behavior represented through identifiable parameters. Tool and material effects are learned directly from differences between machining and dry-run currents rather than separately calibrated cutting-force coefficients. This structure allows the same modeling procedure to be adapted to other robots, cutting tools, and workpiece materials by only updating the relevant robot model and machining parameters and retraining with representative paired data. The evaluator could be integrated into offline programming software to compare candidate trajectories before execution and support energy-aware robot configuration selection.
\end{IEEEpractitioners}


\section{Introduction}
\label{sec:intro}
\IEEEPARstart{R}{obotic} milling enables flexible machining of large and geometrically complex workpieces \cite{Uhlmann2016,verl2019robots}. Long toolpaths and motion constraints can extend machining time \cite{lu2020time}, making energy efficiency an important consideration \cite{Uhlmann2016}. Accurate pre-execution energy prediction can therefore support energy-aware configuration selection and process scheduling \cite{Heredia2021,Gadaleta2021}. However, robotic milling requires the tool center point (TCP) to follow a prescribed path under cutter–workpiece contact. The cutting wrench varies with tool engagement and feed direction, while robot configuration determines its transmission to the joints through the transposed Jacobian. Related studies on pose-dependent cutting-force identification \cite{Hou2023} and structural–force coupling \cite{ChenK2024} highlight the need to account for both process loading and robot configuration when predicting energy.

Existing robot EC models can be grouped into mechanism-based, data-driven, and hybrid approaches. Mechanism-based models estimate joint torques from robot dynamics and calculate electrical demand using actuator and drivetrain loss models.
Wigstrom et al.~\cite{wigstrom2012high} used dynamics-based energy estimation and trajectory time scaling for energy-efficient robot-cell scheduling. Riazi et al.~\cite{riazi2017energy} further optimized energy consumption and peak power while preserving the prescribed robot path and cycle time.
Gadaleta et al. \cite{gadaleta2019optimization} integrated a physical energy model with offline programming to optimize motion parameters and generate executable robot code. Their later experimental study examined how motion settings
affect the EC of a high-payload industrial
robot \cite{Gadaleta2021}.
These models provide a physical description of energy demand, but their accuracy depends on dynamic parameters, friction, drive gains, and electrical losses. Such information is often only partly accessible for commercial robots.

Conventional machining energy models provide a related perspective. Liu et al. \cite{Liu2015} combined cutting power derived from a force model with machine power characteristics in a hybrid milling EC model. Extending this approach to robotic milling requires additional modeling of joint loading and drive behavior. Cutting power at the tool tip alone does not describe how the robot configuration affects individual joint currents and total electrical demand.

Data-driven models reduce dependence on detailed physical identification by learning power or energy from experimental data. Yin et al. \cite{yin2019machine} demonstrated a proof-of-concept approach in which a deep belief network provided the fitness function for
genetic-algorithm trajectory optimization.
Zhang and Yan \cite{zhang2021data} used a backpropagation neural network to estimate EC from robot operating parameters and optimized those parameters using a genetic algorithm.
Later studies used recurrent networks to represent temporal motion patterns. Jiang et al. \cite{jiang2023energy} developed a model based on long short-term memory (LSTM) using joint motion variables for EC prediction and optimization.
Lin et al. \cite{lin2024bn} applied batch normalization to the input-to-hidden transition of an LSTM model to improve convergence and power prediction.
Wang et al. \cite{wang2025novel} combined LSTM with masked multi-head attention and used transfer learning to adapt the model to different operating
conditions. Prediction accuracy nevertheless depends on the coverage of the training data. Some formulations also include measured electrical quantities as inputs \cite{wang2025novel}.
Such quantities require additional estimation for pre-execution use because current and voltage measurements of a candidate trajectory are unavailable before execution.

Hybrid approaches incorporate physical structure into learned
energy models. Yao et al. \cite{yao2024general} separated mechanical power from power losses and developed a network in which an improved bidirectional recurrent neural network (BiRNN) models joint dynamics and a residual network (ResNet) estimates nonlinear
losses.
Wang et al. \cite{wang2025energy} incorporated robot operating mechanisms into
joint-torque and EC prediction, accounting for spatial
dependence between trajectory points and temporal relationships between motion variables and power.
Jiang et al. \cite{jiang2025industrial} employed deep reinforcement learning to identify robot dynamic parameters from trajectory data and incorporated the resulting torque estimates into a neural network-based energy model for EC prediction.
These studies show the value of combining physical knowledge with data-driven estimation. For robotic milling, a central challenge is to predict the configuration-dependent electrical response to cutting from planning-level inputs and to assess transfer to unseen machining geometries and complete toolpath sequences.

The cutting wrench varies along the toolpath, and its contribution to joint loading depends on the instantaneous Jacobian. Li et al.~\cite{li2022general} used deep reinforcement learning to obtain energy-optimal inverse-kinematics solutions for redundant manipulators. In our preliminary experiment (Appendix~\ref{app:preliminary}), task-equivalent trajectories with identical TCP paths and cutting parameters differed in robot energy consumption by up to 8.1\%. This variation motivates energy-aware configuration selection and creates a need to estimate candidate trajectory energy before physical execution. The evaluator must also be tested on machining geometries and complete trajectories that are
excluded from model development.

Using milling with a commonly used flat-end mill as an example, this paper proposes a Jacobian-aware, current-first cascade model for pre-execution robot energy prediction in milling.
The main contributions are as follows:
\begin{enumerate}
    \item A novel Jacobian-aware, current-first cascade is developed for pre-execution robot energy prediction in milling, with a cutting-induced current (CIC) stage that explicitly predicts the configuration-dependent joint-current response to time-varying cutting loads.

    \item SK-RE pairs simulated kinematics with measured electrical targets, enabling pre-execution robot energy prediction using only simulated kinematics and planned machining parameters.

    \item Unseen-feature tests assess transfer to different machining geometries and complete toolpath sequences. Ablations and reference-model comparisons examine the contributions of the current stages, simulation-derived training inputs, and training-data coverage.
\end{enumerate}

Section~\ref{sec:method} presents the methodology.
Section~\ref{sec:exp} describes the experiments and data
construction. Section~\ref{sec:results} reports the prediction results, ablations, and reference-model comparisons. Sections~\ref{sec:discussion} and~\ref{sec:conclusion} provide the discussion and conclusion.
\section{Methodology}
\label{sec:method}

\subsection{Power Composition and the Current-First Cascade}

Let $s_t$ denote sample $t$ of a planned joint-space trajectory. Robot electrical power consists of drive power $P_{\mathrm{drive}}$, approximately constant control power $P_{\mathrm{c}}$, and internal auxiliary consumption $P_{\mathrm{a}}$ under a fixed
setup~\cite{yao2024general}:
\begin{equation}
P_{\mathrm{robot}}(s_t)
=
P_{\mathrm{drive}}(s_t)+P_{\mathrm{c}}+P_{\mathrm{a}}.
\label{eq:power_decomposition}
\end{equation}
Here, $P_{\mathrm{a}}$ refers to auxiliary consumption within the robot system; spindle and machining auxiliary equipment are excluded. The joint mechanical power is
\begin{equation}
P_{\mathrm{mc}}(s_t)
=
\bm{\tau}^{\top}(s_t)\dot{\bm{q}}(s_t).
\label{eq:mechanical_power}
\end{equation}
Joint torque is related to motor current through
\begin{equation}
\bm{\tau}(s_t)=\mathbf{G}\mathbf{i}(s_t),
\qquad
\mathbf{G}
=
\mathrm{diag}(k_{t,1}g_1,\ldots,k_{t,n}g_n).
\label{eq:geargain}
\end{equation}
Here, $\mathbf{G}$ contains the effective gear gains.
The parameters $k_{t,j}$ and $g_j$ denote the motor torque constant and gear ratio of joint $j$, respectively.

\begin{figure*}[!t]
  \centering
  \includegraphics[width=\textwidth]{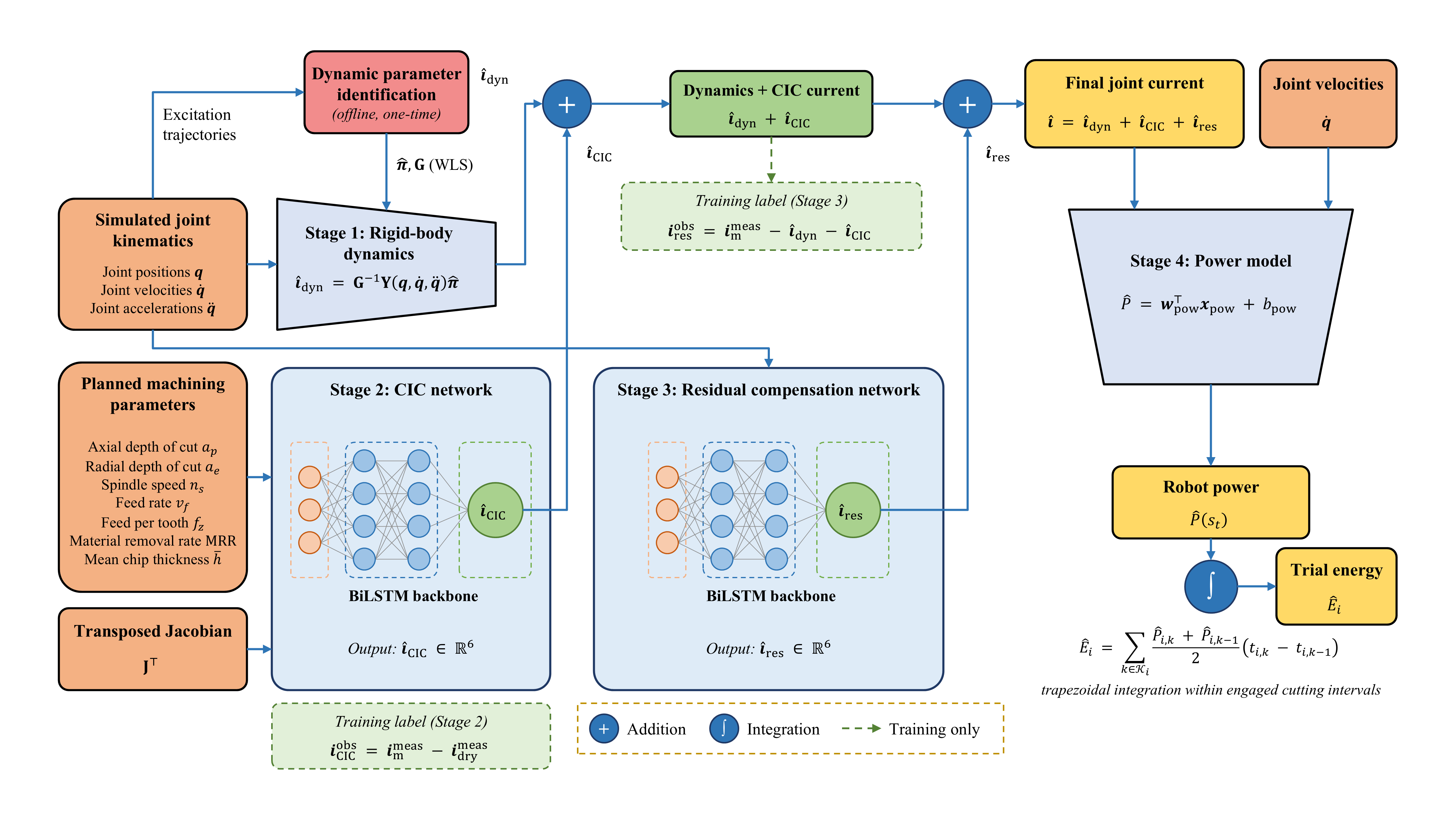}
 \caption{Jacobian-aware current-first cascade for pre-execution robot
energy prediction. Simulated kinematics and planned machining
parameters drive the current stages, followed by power estimation
and energy integration.}
  \label{fig:framework}
\end{figure*}

Joint current, therefore, provides an intermediate representation of
motion- and contact-related loading. It also supplies features relevant
to electrical losses that rigid-body dynamics alone do not describe~\cite{yao2024general,wang2025energy}. The proposed cascade predicts joint current from planning-level inputs and then estimates
power from the predicted current and joint velocity.
Figure~\ref{fig:framework} summarizes the four stages. Identification is performed once offline.

\subsection{Stage 1: Rigid-Body Dynamic Model and Parameter Identification}
Stage~1 provides a physics-based current estimate for the subsequent learnable stages. For an $n$-degree-of-freedom (DOF) serial manipulator executing the interpolated path $s_t$, the joint torque follows the rigid-body Newton-Euler dynamic equation ~\cite{khalil2011dynamic}
\begin{equation}
  \begin{aligned}
  \bm{\tau}(s_t) ={}&\mathbf{M}(\bm{q})\ddot{\bm{q}}
  +\mathbf{C}(\bm{q},\dot{\bm{q}})\dot{\bm{q}}
  +\bm{g}(\bm{q})\\
  &+\bm{\tau}_\mathrm{f}(\dot{\bm{q}})
  +\mathbf{J}(\bm{q})^{\top}\mathbf{W}_\text{cut}(s_t).
  \end{aligned}
  \label{eq:rbd}
\end{equation}
Here $\mathbf{M}$ is the inertia matrix, $\mathbf{C}(\bm{q},\dot{\bm{q}})\dot{\bm{q}}$ contains Coriolis and centrifugal torques, and $\bm{g}$ is the gravitational torque. Joint friction is modeled as
\begin{equation}
\bm{\tau}_\mathrm{f}(\dot{\bm{q}})
=
\mathbf{F}_\mathrm{c}\operatorname{sgn}(\dot{\bm{q}})
+
\mathbf{F}_\mathrm{v}\dot{\bm{q}}
+
\bm{f}_\mathrm{o}.
\end{equation}
The term $\mathbf{W}_\text{cut}(s_t)$ denotes the cutting wrench at the TCP. For dry-run trajectories, where $\mathbf{W}_\text{cut}=\bm{0}$,
Eq.~\eqref{eq:rbd} is linear in the dynamic parameters and can be written in regressor form~\cite{gautier1997dynamic}
\begin{equation}
  \bm{\tau}(s_t)=\mathbf{Y}(\bm{q},\dot{\bm{q}},\ddot{\bm{q}})\,\bm{\pi},
  \label{eq:regressor}
\end{equation}
where $\mathbf{Y}$ is the regressor matrix and $\bm{\pi}$ stacks the link-wise standard parameter vectors
\begin{equation}
\begin{aligned}
\bm{\pi}_j = [&I_{xx},  I_{xy}, I_{xz}, I_{yy}, I_{yz}, I_{zz},\\
& H_x, H_y, H_z, m, J_\mathrm{m}, f_\mathrm{v}, f_\mathrm{c}, f_\mathrm{o}]_j^{\top}\in\mathbb{R}^{14}
\end{aligned}
\end{equation}
which comprises six independent inertia-tensor entries, three first-order mass moments, the link mass, the rotor inertia, and the viscous, Coulomb, and offset friction coefficients.  Linearly dependent entries are regrouped into a base parameter set. The base parameters $\hat{\bm{\pi}}$ and the gear-gain matrix $\mathbf{G}$ are identified jointly by weighted least squares on persistently exciting periodic
trajectories~\cite{swevers1997optimal,gautier2014global,mamedov2020practical}. Inverting Eq.~\eqref{eq:geargain}
then yields the preliminary  current
\begin{equation}
  \hat{\mathbf{i}}_\text{dyn}(s_t)
  =\mathbf{G}^{-1}\mathbf{Y}(\bm{q},\dot{\bm{q}},\ddot{\bm{q}})\,
  \hat{\bm{\pi}}.
  \label{eq:idyn}
\end{equation}
Stage~1 contains no trainable neural parameters and therefore serves as a physics-informed basis for subsequent modules. During machining, the interaction term $\mathbf{J}(\bm{q})^{\top}\mathbf{W}_\text{cut}$ in Eq.~\eqref{eq:rbd} no longer vanishes. Its mechanistic construction and the resulting CIC input set are developed in the next subsection.

\subsection{Cutting-Force Modeling and the CIC Input Set}
For a flat-end mill of radius $R$ with $N_f$ flutes, the differential cutting force on an infinitesimal cutting-edge element consists of a shear component proportional to the instantaneous chip thickness
$h(\phi)=f_z\sin\phi$ and an edge component associated with rubbing and plowing~\cite{budak1996prediction}:
\begin{equation}
  dF_q(\phi,z)=K_{qe}\,dS+K_{qc}\,f_z\sin\phi\,dz,
  \qquad q\in\{t,r,a\},
  \label{eq:altintas}
\end{equation}
where $f_z$ is the feed per tooth, $\phi$ the immersion angle of the edge element at axial position $z$, $dS$ the differential edge length, and $K_{qc}$ (N/mm$^2$), $K_{qe}$ (N/mm) the shear- and edge-specific coefficients of the tool-workpiece pair. The differential force components are integrated over the engaged axial depth and immersion arcs of all flutes to obtain the TCP cutting
wrench, $\mathbf{W}_\text{cut}(\phi)=[\mathbf{F}_\text{cut}^\top,
\mathbf{M}_\text{cut}^\top]^\top$.
By the principle of virtual work, \begin{equation}
\bm{\tau}_\mathrm{cic}
=
\mathbf{J}(\bm{q})^{\top}\mathbf{W}_\mathrm{cut},
\end{equation}
and the corresponding cutting-induced current is
\begin{equation}
  \mathbf{i}_\text{cic}
  =\mathbf{G}^{-1}\bm{\tau}_\text{cic}
  =\mathbf{G}^{-1}\mathbf{J}(\bm{q})^\top{\mathbf{W}}_\text{cut}.
  \label{eq:cic}
\end{equation}

The CIC network uses planned machining parameters and the transposed Jacobian to estimate contact-related current without explicit cutting-coefficient calibration for the tested tool-workpiece pair. The machining-parameter vector is
\begin{equation}
\bm{p}_m
=
[a_p,a_e,n_s,v_f,f_z,\mathrm{MRR},\bar{h}]^{\top}
\in\mathbb{R}^{7}.
\label{eq:machining_parameters}
\end{equation}
The tool diameter and flute count are fixed in these experiments.
For nominal radial engagement, the mean chip thickness is
\begin{equation}
\bar{h}
=
\frac{a_e f_z}{R\phi_e},
\qquad
\phi_e
=
\arccos\left(1-\frac{a_e}{R}\right),
\label{eq:mean_chip_thickness}
\end{equation}
where $0<a_e\leq2R$. With $v_f$ in mm/s and $n_s$ in rpm,
$f_z=60v_f/(n_sN_f)$ and $\mathrm{MRR}=a_pa_ev_f$.
The flattened transposed Jacobian, $\mathrm{vec}(\mathbf{J}^\top)\in\mathbb{R}^{36}$, encodes the configuration-dependent mapping from TCP load to joint load. This input design avoids explicit cutting-coefficient calibration for the tested tool-workpiece pair.

\begin{figure}[!tb]
    \centering
    \includegraphics[width=1\columnwidth]{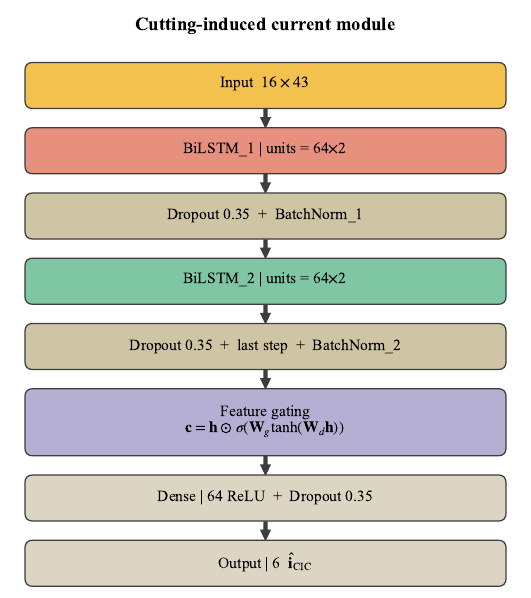}
    \caption{CIC network with a 16-sample input window containing seven
machining parameters and the 36-element transposed Jacobian.}
    \label{fig:cic_network}
\end{figure}

\begin{figure}[!tb]
    \centering
    \includegraphics[width=1\columnwidth]{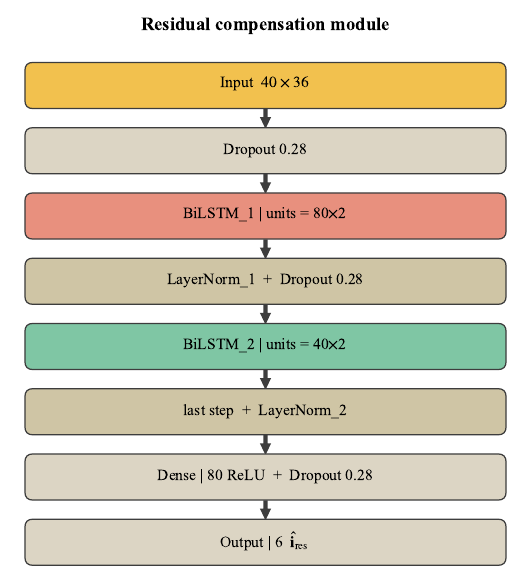}
    \caption{Residual-current network with a 40-sample input window
containing joint kinematics and their nonlinear terms.}
    \label{fig:res_network}
\end{figure}

The CIC and residual-current networks use bidirectional LSTMs
(BiLSTMs) to encode temporal context within each input window~\cite{hochreiter1997long,schuster1997bidirectional}.
The power stage uses a linear map. Their architectures are shown in Figs. \ref{fig:cic_network}-\ref{fig:power_network}.

\subsection{Stage 2: Cutting-Induced Current Network}

The CIC network predicts contact-related joint current from machining parameters and the transposed Jacobian. Its sample-level input is
\begin{equation}
\mathbf{x}_{\mathrm{CIC}}(s_t)
=
\left[
\bm{p}_m^{\top}(s_t),
\mathrm{vec}\left(\mathbf{J}^{\top}(\bm{q}(s_t))\right)^{\top}
\right]^{\top}
\in\mathbb{R}^{43}.
\label{eq:xcic}
\end{equation}
The temporal input contains $w_{\mathrm{CIC}}=16$ samples:
\begin{equation}
\mathbf{X}_{\mathrm{CIC}}(s_t)
=
\begin{bmatrix}
\mathbf{x}_{\mathrm{CIC}}^{\top}(s_{t-w_{\mathrm{CIC}}+1})\\
\vdots\\
\mathbf{x}_{\mathrm{CIC}}^{\top}(s_t)
\end{bmatrix}
\in\mathbb{R}^{16\times43}.
\label{eq:Xcic}
\end{equation}
The CIC target is the joint-current difference between machining and
dry runs of the same programmed toolpath. Both current streams are
aligned to the simulation trajectory before subtraction:
\begin{equation}
\mathbf{i}^{\mathrm{obs}}_{\mathrm{CIC}}(s_t)
=
\mathbf{i}^{\mathrm{meas}}_{\mathrm{m}}(s_t)
-
\mathbf{i}^{\mathrm{meas}}_{\mathrm{dry}}(s_t)
\in\mathbb{R}^{6},
\label{eq:cic_target}
\end{equation}
where the two measured current vectors correspond to the same
simulation sample $s_t$. The CIC network learns the mapping
\begin{equation}
\hat{\mathbf{i}}_{\mathrm{CIC}}(s_t)
=
f_{\mathrm{CIC}}
\left(\mathbf{X}_{\mathrm{CIC}}(s_t)\right).
\label{eq:cic_mapping}
\end{equation}

The CIC estimator uses two BiLSTM layers with 64 hidden units per
direction, dropout of 0.35, and batch
normalization~\cite{ioffe2015batch,lin2024bn}.
The last-step representation after the second normalization layer,
$\mathbf{h}\in\mathbb{R}^{128}$, is passed through a feature gate:
\begin{equation}
\mathbf{c}
=
\mathbf{h}\odot
\sigma\left(
\mathbf{W}_g\tanh(\mathbf{W}_d\mathbf{h})
\right),
\label{eq:gate}
\end{equation}
where $\sigma$ is the sigmoid function and $\odot$ denotes element-wise
multiplication. The gated features pass through a 64-unit ReLU head
with dropout and a six-current output layer. 

\subsection{Stage 3: Residual Compensation Network}

The dynamic and CIC estimates may retain a systematic joint-current error from unmodeled friction, drive behavior, and measurement alignment. A residual network estimates this remaining current component.

The residual network uses $w_{\mathrm{res}}=40$ samples of joint
kinematics and their nonlinear terms:
\begin{equation}
\begin{aligned}
\mathbf{x}_{\mathrm{res}}(s_t)
=
[&\bm{q}^{\top},
\dot{\bm{q}}^{\top},
\ddot{\bm{q}}^{\top},
(\dot{\bm{q}}^{\circ2})^{\top},\\
&(\ddot{\bm{q}}^{\circ2})^{\top},
(\dot{\bm{q}}\odot\ddot{\bm{q}})^{\top}]^{\top}(s_t)
\in\mathbb{R}^{36},
\end{aligned}
\label{eq:xres}
\end{equation}
where $(\cdot)^{\circ2}$ denotes element-wise squaring.
The temporal input is
\begin{equation}
\mathbf{X}_{\mathrm{res}}(s_t)
=
\begin{bmatrix}
\mathbf{x}_{\mathrm{res}}^{\top}(s_{t-w_{\mathrm{res}}+1})\\
\vdots\\
\mathbf{x}_{\mathrm{res}}^{\top}(s_t)
\end{bmatrix}
\in\mathbb{R}^{40\times36}.
\label{eq:Xres}
\end{equation}
The observed residual target is defined as the difference between the measured machining current and
the two upstream current estimates:
\begin{equation}
  \mathbf{i}^{\mathrm{obs}}_{\mathrm{res}}(s_t)
  =
  \mathbf{i}^{\mathrm{meas}}_{\mathrm{m}}(s_t)
  -
  \hat{\mathbf{i}}_{\mathrm{dyn}}(s_t)
  -
  \hat{\mathbf{i}}_{\mathrm{CIC}}(s_t)
  \in \mathbb{R}^{6}.
  \label{eq:res_target}
\end{equation}
The residual network then learns
\begin{equation}
  \hat{\mathbf{i}}_{\mathrm{res}}(s_t)
  =
  f_{\mathrm{res}}
  \left(
  \mathbf{X}_{\mathrm{res}}(s_t)
  \right),
  \label{eq:res_mapping}
\end{equation}
where $\hat{\mathbf{i}}_{\mathrm{res}}$ is the current component not captured by the dynamic and CIC stages.

The residual estimator uses two BiLSTM layers with 80 and 40 hidden units per direction, layer normalization, and dropout of 0.28. The last-step representation passes through an 80-unit ReLU head and a six-current output layer. The complete joint current prediction is obtained by summing the three current components:
\begin{equation}
  \hat{\mathbf{i}}(s_t)
  =
  \hat{\mathbf{i}}_{\mathrm{dyn}}(s_t)
  +
  \hat{\mathbf{i}}_{\mathrm{CIC}}(s_t)
  +
  \hat{\mathbf{i}}_{\mathrm{res}}(s_t).
  \label{eq:current_sum}
\end{equation}

\subsection{Stage 4: Power Model}
\label{sec:power_model}

The final stage maps the predicted joint current in Eq.~\eqref{eq:current_sum} to the instantaneous electrical power drawn by the robot. Total robot power is not determined by mechanical joint power alone. It also includes current-dependent copper loss, speed-dependent drive losses, and quasi-stationary control consumption. The power model therefore uses an expanded feature set constructed from the joint velocity and the predicted current.

For sample $s_t$, the feature vector is
\begin{equation}
\begin{aligned}
\mathbf{x}_{\mathrm{pow}}(s_t)
=
[&\dot{\bm{q}}^{\top},
(\dot{\bm{q}}^{\circ2})^{\top},
(\dot{\bm{q}}^{\circ3})^{\top},\\
&(\hat{\mathbf{i}}^{\circ2})^{\top},
\hat{\mathbf{i}}^{\top},
(\hat{\mathbf{i}}\odot\dot{\bm{q}})^{\top}]^{\top}(s_t)
\in\mathbb{R}^{36}.
\end{aligned}
\label{eq:xpow}
\end{equation}
The velocity terms describe speed-dependent behavior, while
$\hat{\mathbf{i}}^{\circ2}$ provides the current-squared features associated with copper loss. The products $\hat{\mathbf{i}}\odot\dot{\bm{q}}$ provide joint-level electromechanical power features; their gain-weighted sum gives the mechanical power estimate under Eq.~\eqref{eq:geargain}. The bias represents approximately constant consumption within the robot system.

\begin{figure}[!tb]
    \centering
    \includegraphics[width=1\columnwidth]{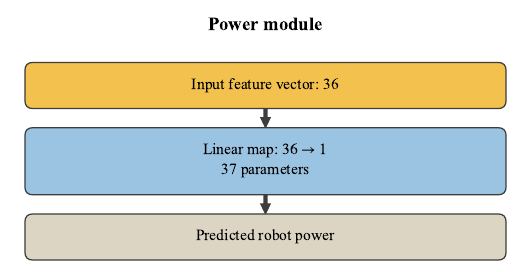}
    \caption{Power model. The 36-dimensional final feature vector is mapped to instantaneous robot power.}
    \label{fig:power_network}
\end{figure}

Power is predicted by a sample-level linear map:
\begin{equation}
\hat{P}(s_t)
=
\mathbf{w}_{\mathrm{pow}}^{\top}
\mathbf{x}_{\mathrm{pow}}(s_t)
+b_{\mathrm{pow}},
\label{eq:pow_mapping}
\end{equation}
where $\mathbf{w}_{\mathrm{pow}}\in\mathbb{R}^{36}$ and $b_{\mathrm{pow}}\in\mathbb{R}$. The model contains 37 trainable parameters, as shown in Figure~\ref{fig:power_network}.

The predicted energy of trial $i$ is
\begin{equation}
\hat{E}_i
=
\sum_{k\in\mathcal{K}_i}
\frac{\hat{P}_{i,k}+\hat{P}_{i,k-1}}{2}
\left(t_{i,k}-t_{i,k-1}\right),
\label{eq:energy_integration}
\end{equation}
where $\mathcal{K}_i$ contains adjacent sample pairs within the same
engaged cutting interval.

\subsection{Evaluation Metrics}
\label{sec:evaluation_metrics}

Let $y_k$ and $\hat{y}_k$ denote the measured and predicted values of a current or power signal, respectively, and let $M$ be the number of evaluated samples. The root mean squared error (RMSE), mean absolute error (MAE), and coefficient of determination ($R^2$) are defined as:
\begin{align}
\mathrm{RMSE}
&=
\sqrt{\frac{1}{M}\sum_{k=1}^{M}(\hat{y}_k-y_k)^2},
\label{eq:rmse}\\
\mathrm{MAE}
&=
\frac{1}{M}\sum_{k=1}^{M}|\hat{y}_k-y_k|,
\label{eq:mae}\\
\mathrm{R^2}
&=
1-
\frac{\sum_{k=1}^{M}(\hat{y}_k-y_k)^2}
{\sum_{k=1}^{M}(y_k-\bar{y})^2},
\label{eq:r2}
\end{align}
where $\bar{y}$ is the mean of the measured signal. Joint-current accuracy is reported using RMSE, MAE, and $R^2$, while instantaneous-power accuracy is reported using RMSE and $R^2$. For each evaluation set or machining feature, these metrics are calculated over concatenated samples.

Cumulative energy is evaluated separately for each trial. For a set containing
$N$ trials, the absolute relative energy error of trial $i$ is
\begin{equation}
\delta_{E,i}
=
\frac{|\hat{E}_i-E_i|}{E_i}\times100\%,
\label{eq:trial_energy_error}
\end{equation}
where $E_i$ and $\hat{E}_i$ are the measured and predicted cutting-phase
energies. The reported energy result is the mean and sample standard deviation
across trials:
\begin{align}
\bar{\delta}_E
&=
\frac{1}{N}\sum_{i=1}^{N}\delta_{E,i},
\label{eq:mean_energy_error}\\
s_E
&=
\sqrt{
\frac{1}{N-1}
\sum_{i=1}^{N}
(\delta_{E,i}-\bar{\delta}_E)^2
}.
\label{eq:std_energy_error}
\end{align}
Subsequent tables and figures report this quantity as
$\delta_E=\bar{\delta}_E\pm s_E$.

\section{Experimental Setup}
\label{sec:exp}

\subsection{Robotic Machining Platform}
\begin{figure*}[!t]
    \centering
    \includegraphics[width=1\textwidth]{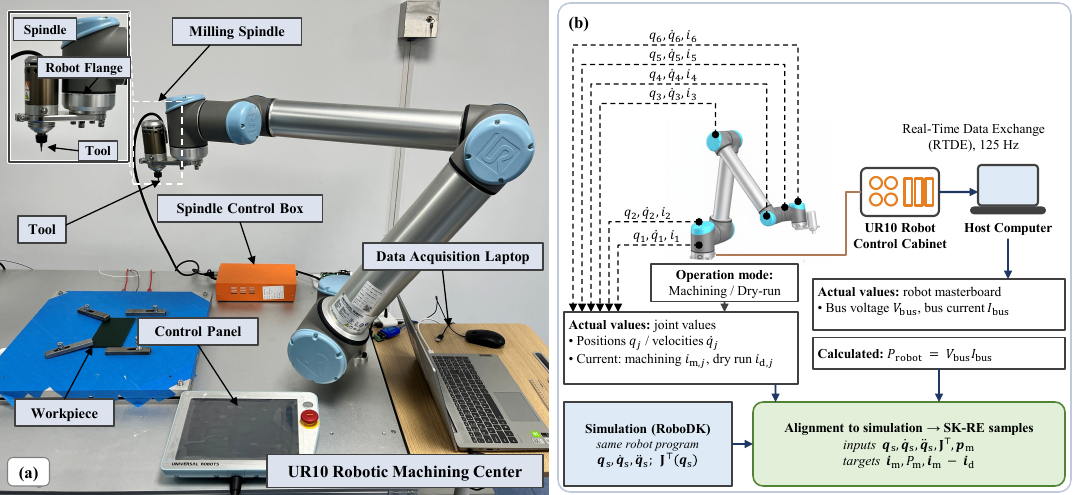}
    \caption{Experimental platform and data acquisition. (a) UR10 robotic milling cell. (b) RTDE-based acquisition of robot kinematic and electrical signals.}
    \label{fig:expdata}
\end{figure*}
Figure~\ref{fig:expdata}(a) shows the milling platform: a UR10 manipulator, a STEPCRAFT HF-500 v2 spindle mounted on its flange, a workpiece fixture, and a data-acquisition computer. A 3~mm two-flute flat-end mill was used in all experiments.

Figure~\ref{fig:expdata}(b) shows the data-acquisition scheme. Joint positions $\bm{q}$, velocities $\dot{\bm{q}}$, and currents $\mathbf{i}$, together with bus voltage $V_{\mathrm{bus}}$ and current $I_{\mathrm{bus}}$, were acquired from the UR10 controller through the Real-Time Data Exchange (RTDE) interface at 125~Hz. Robot power $P_{\mathrm{robot}}$ was calculated as

\begin{equation}
P_{\mathrm{robot}}(t)
=
V_{\mathrm{bus}}(t)I_{\mathrm{bus}}(t).
\label{eq:robot_power_measurement}
\end{equation}
$V_{\mathrm{bus}}$ remained nearly constant, with a mean value of 48.25~V.

A $150\times90\times5$~mm machining-wax workpiece was used to reduce cutting-induced deflection and improve experimental repeatability. The wax trials provide a controlled, low-load evaluation of the SK-RE pipeline and energy model. Robot energy in this study includes only the electrical energy drawn by the manipulator and its drive and control system. Spindle and machining auxiliary equipment energy are excluded.

\subsection{Machining Features and Directional Diversity}
\label{sec:features}

The dataset contains four machining-feature families with different geometries, feed-direction patterns, and robot postures:
\begin{itemize}
    \item \textit{One-way slot} (OW): a 150~mm straight slot used as the baseline feature;
    \item \textit{U-slot} (US): a semicircular pocket machined at four workpiece locations, U1-U4;
    \item \textit{P1}: a cross-obround pocket formed by two overlapping obround slots;
    \item \textit{P2}: an octagonal pocket containing a central island and two circular islands.
\end{itemize}

\begin{figure}[!tb]
    \centering
    \includegraphics[width=1\columnwidth]{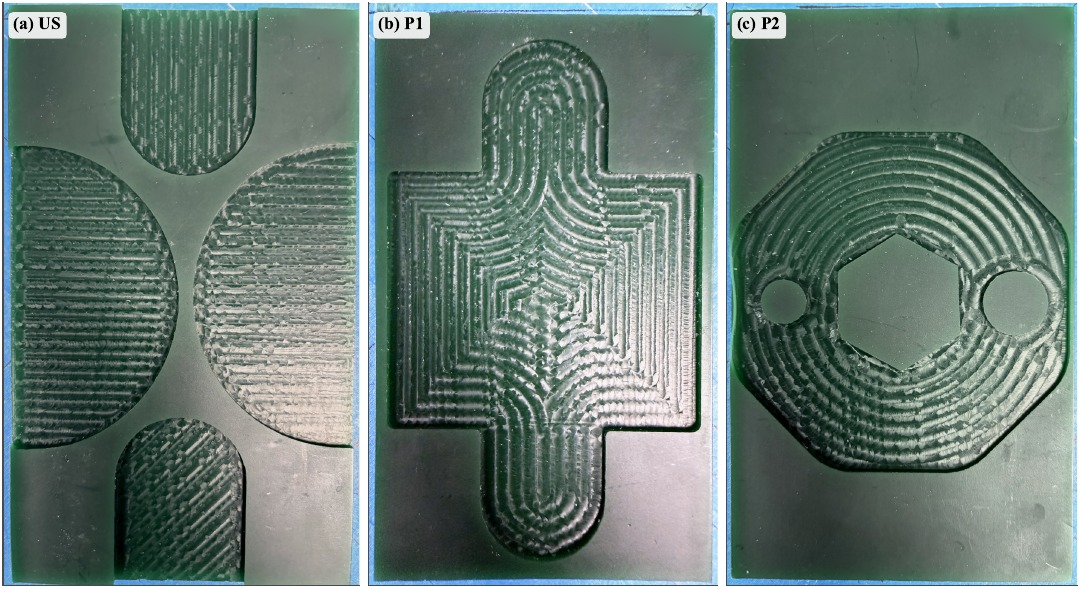}
    \caption{Machined wax features: (a) U-slot (US) at four workpiece locations, (b) cross-obround pocket (P1), and (c) octagonal pocket (P2) with a central island and two circular islands.}
    \label{fig:workpieces}
\end{figure}

\begin{figure*}[!tb]
    \centering
    \includegraphics[width=1\textwidth]{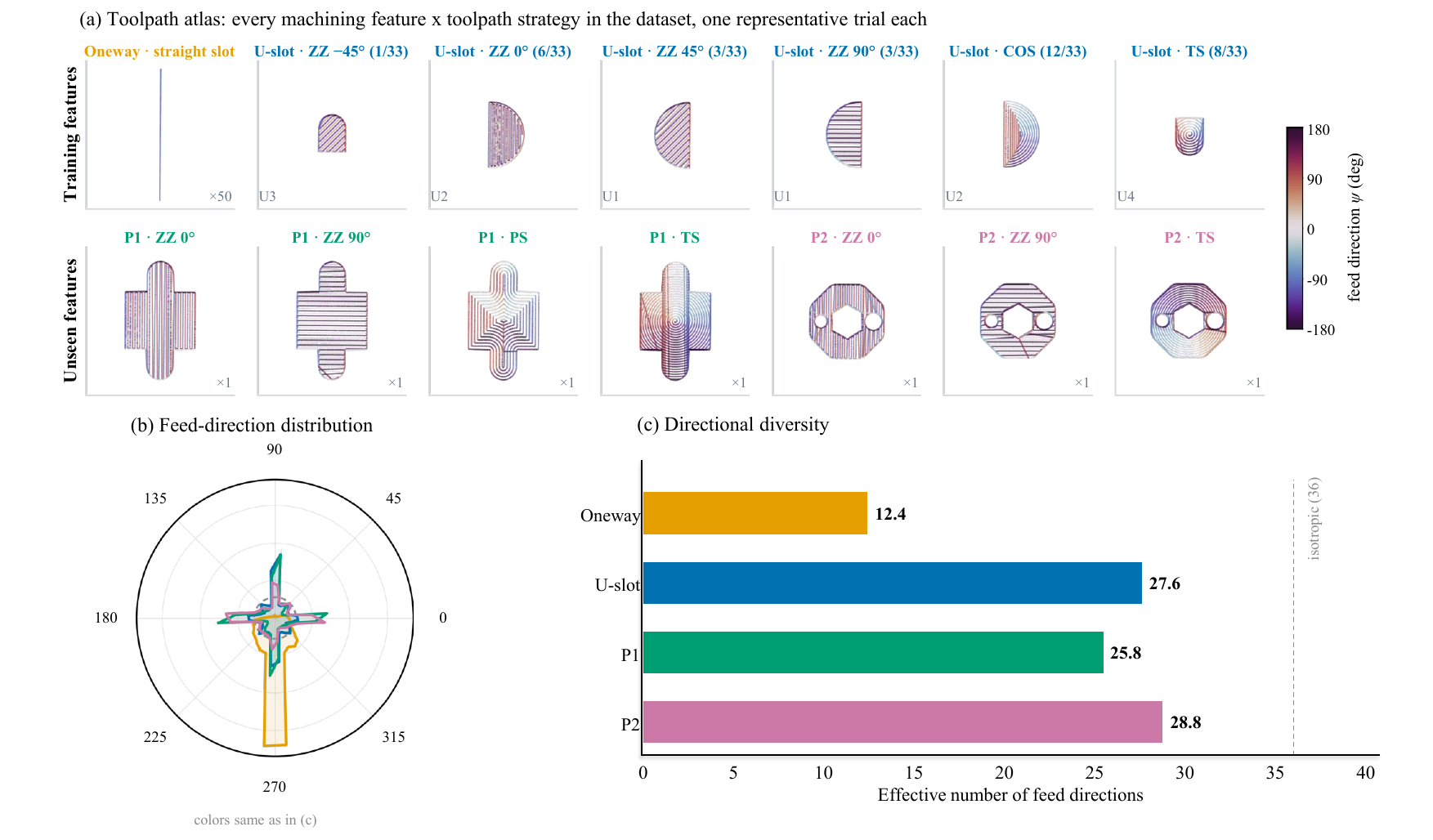}
   \caption{Dataset coverage and feed-direction diversity. (a) Representative toolpath for each feature-strategy combination. Path color denotes feed direction, and the numbers indicate trial counts. (b) Speed-weighted feed-direction distributions. (c) Directional diversity expressed as the effective number of feed directions. The dashed line denotes uniform coverage over all 36 sectors.}
    \label{fig:features}
\end{figure*}

Figure~\ref{fig:workpieces} shows the machined US, P1, and P2 features. The OW and US trials constitute the model-development set, while all P1 and P2 trials form the unseen-feature test set. The toolpath strategies include zigzag paths (ZZ) at different orientations, a constant-overlap spiral (COS), a parallel spiral (PS), and a true spiral (TS). Figure~\ref{fig:features}(a) summarizes the feature-strategy combinations and their trial counts.

Following standard circular-data analysis~\cite{mardia2000directional}, the in-plane TCP feed direction was divided into 36 sectors of $10^{\circ}$. Each sample was weighted by TCP feed speed, so Figure~\ref{fig:features}(b) represents the proportion of the engaged path traveled in each direction.

Directional diversity was quantified by the inverse-Simpson effective number, equivalent to the order-two Hill number~\cite{hill1973diversity}. The resulting $N_{\mathrm{eff}}$ is the equivalent number of equally represented direction sectors. OW has the lowest directional diversity, with $N_{\mathrm{eff}}=12.4$, whereas US, P1, and P2 range from 25.8 to 28.8. Thus, the model-development set includes both directionally concentrated and multidirectional toolpaths, while P1 and P2 test generalization to unseen feature geometries and feed-direction sequences.
\subsection{SK-RE Dataset Construction}
\label{sec:skre}

Each prescribed toolpath produces three data streams. The real machining run ($m$) records joint position $\mathbf{q}_m$, joint current $\mathbf{i}_m$, and robot power $P_m$ during cutting. The real dry run ($d$) executes the same robot program without material removal and records $\mathbf{q}_d$ and $\mathbf{i}_d$. The simulation run ($s$) executes the corresponding RoboDK program and provides $\mathbf{q}_s$, $\dot{\mathbf{q}}_s$, and $\ddot{\mathbf{q}}_s$.

Differences in local trajectory progression require each real stream
to be aligned independently to the simulation sequence. Related
sequence-alignment methods preserve temporal
order~\cite{macrae2010accurate,lacoquelle2022deviation}.
Here, alignment uses a forward-constrained nearest-neighbor search
in joint space.

Let $\mathbf{q}_{s,i}$ denote simulation sample $i$, and let $\mathbf{q}_{r,j}$ denote sample $j$ from real stream $r\in\{m,d\}$. Their joint-space distance is
\begin{equation}
d_{i,j}^{(r)}
=
\frac{1}{\sqrt{6}}
\left\|
\mathbf{q}_{s,i}-\mathbf{q}_{r,j}
\right\|_2.
\label{eq:alignment_distance}
\end{equation}

For each simulation sample, the search is limited to the next $W=50$ real samples after the previous match. The matched index is
\begin{equation}
\alpha_r(i)
=
\operatorname*{arg\,min}_{\alpha_r(i-1)<j\leq J_i^{(r)}}
d_{i,j}^{(r)},
\label{eq:alignment_index}
\end{equation}
where $J_i^{(r)}=\min\{\alpha_r(i-1)+W,N_r\}$, $N_r$ is the length of real stream $r$, and $\alpha_r(0)=0$. The forward constraint preserves the toolpath order and prevents backward matching between repeated passes.

After alignment, each simulation sample is paired with the corresponding physical measurements:
\begin{equation}
\begin{aligned}
&\left(
\mathbf{q}_{s,i},
\dot{\mathbf{q}}_{s,i},
\ddot{\mathbf{q}}_{s,i}
\right)
\longrightarrow
\left(
\mathbf{i}_{m,\alpha_m(i)},
P_{m,\alpha_m(i)},
\mathbf{i}_{d,\alpha_d(i)}
\right).
\end{aligned}
\label{eq:skre_mapping}
\end{equation}

The machining and dry-run currents are matched independently to
the same simulation sample through $\alpha_m(i)$ and $\alpha_d(i)$.
Their difference provides the CIC target defined in
Eq.~\eqref{eq:cic_target}. Simulation-derived kinematics supply the
model inputs, while aligned real-machine measurements provide
electrical supervision.

\subsection{Dataset Partition and Evaluation Protocol}
\label{sec:data_partition}

The dataset contains 90 trials and 1,896,568 synchronized cutting samples, corresponding to 252.9~min of engaged motion. Each trial contains one simulation trajectory and the corresponding machining and dry runs. The experimental conditions and dataset composition are summarized in Table~\ref{tab:machining_matrix}.

The 50 OW trials and 33 US trials form the model-development set. These 83 trials are divided using a group-wise 80/20 train/within-feature test split, with each complete trial treated as one group. No samples from a within-feature test trial are used for training..

All four P1 trials and three P2 trials form the unseen-feature test set. These seven trials contain 602,630 samples and 80.4~min of engaged machining and are excluded from training, validation, hyperparameter tuning, and architecture selection. Although some toolpath families are shared with the model-development set, no P1 or P2 geometry or complete trajectory appears in training.

\begin{table}[!b]
    \caption{Experimental conditions and SK-RE dataset composition.}
    \label{tab:machining_matrix}
    \centering
    \footnotesize
    \renewcommand{\arraystretch}{1.08}
    \setlength{\tabcolsep}{4pt}
    \begin{tabular}{@{}lp{0.60\columnwidth}@{}}
        \hline
        \textbf{Item} & \textbf{Setting} \\
        \hline
        Manipulator & Universal Robots UR10 \\
        Spindle & STEPCRAFT HF-500 v2 \\
        Cutting tool & 3~mm two-flute flat-end mill \\
        Workpiece & Machining wax, $150\times90\times5$~mm \\
        Sampling rate & 125~Hz \\
        Feed rate & 1.0-8~mm/s\\
        Axial depth of cut & 0.5-2.0~mm\\
        Spindle speed & 1200-5000~rpm \\
        Machining features & OW, US, P1, and P2 \\
        Toolpath strategies & ZZ, COS, PS, and TS \\
        Development set & 50 OW and 33 US trials \\
        Development split & 993,605 training and 300,333 within-feature test samples \\
        Unseen-feature test set & 4 P1 and 3 P2 trials; 602,630 samples \\
        Total dataset & 90 trials; 1,896,568 samples\\
        \hline
    \end{tabular}
\end{table}

\section{Results}
\label{sec:results} 

Results are reported for the within-feature and the unseen-feature test sets using the protocol in Section~\ref{sec:data_partition}. 

\subsection{Stagewise Joint-Current Prediction}
\label{sec:current_results}

\begin{figure*}[!t]
    \centering
    \includegraphics[width=1\textwidth]{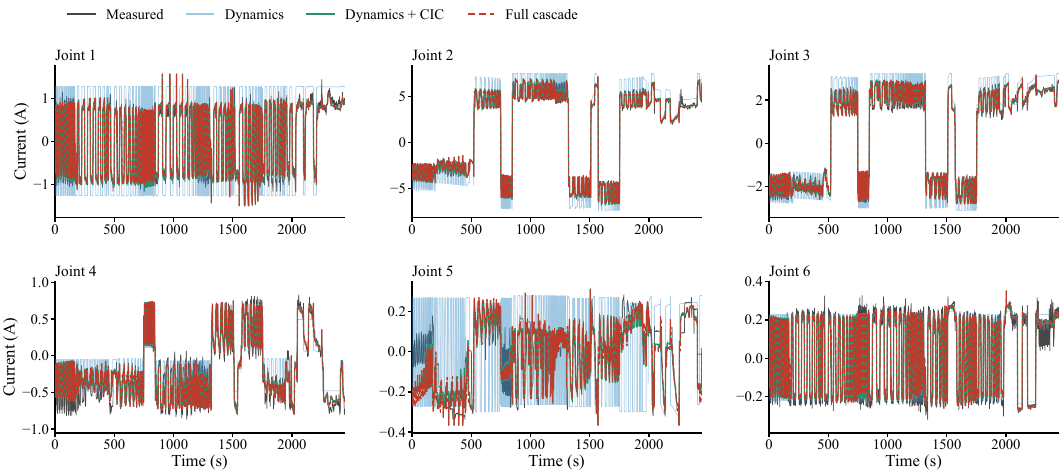}
    \caption{Stagewise joint-current prediction on the within-feature test set. Black: measured current; blue: rigid-body dynamics; green: dynamics plus CIC; red: full cascade.}
    \label{fig:current_stages}
\end{figure*}

Figure~\ref{fig:current_stages} compares measured joint currents with the
outputs of the rigid-body dynamics stage, the dynamics-plus-CIC stages, and
the full cascade. The dynamics stage captures the main
configuration-dependent current levels but retains systematic errors during
cutting. The CIC stage provides the main correction for J1-J5. The residual
stage produces smaller changes, and its output largely overlaps the
dynamics-plus-CIC prediction. For J6, the predictions follow the main current reversals, while some smaller fluctuations remain unresolved.

Table~\ref{tab:current_internal_test} reports per-joint accuracy. The three $R^2$ columns show the successive current estimates, while RMSE and MAE refer to the full cascade. CIC substantially improves $R^2$ for J1-J5. The residual stage has a smaller effect on these pointwise current metrics.
\begin{table}[t]
    \caption{Joint-current accuracy on the within-feature test set.}
    \label{tab:current_internal_test}
    \centering
    \footnotesize
    \renewcommand{\arraystretch}{1.12}
    \setlength{\tabcolsep}{2.5pt}

    \begin{tabular}{cccccc}
        \hline
        \textbf{Joint} &
        \shortstack{\textbf{Dynamics}\\\textbf{$R^2$}} &
        \shortstack{\textbf{Dynamics+CIC}\\\textbf{$R^2$}} &
        \shortstack{\textbf{Full}\\\textbf{$R^2$}} &
        \shortstack{\textbf{Full}\\\textbf{RMSE (A)}} &
        \shortstack{\textbf{Full}\\\textbf{MAE (A)}} \\
        \hline
        J1 & 0.61 & 0.91 & 0.91 & 0.24 & 0.18 \\
        J2 & 0.94 & 0.98& 0.99 & 0.39 & 0.25 \\
        J3 & 0.92 & 0.97& 0.99 & 0.21 & 0.15 \\
        J4 & 0.84 & 0.90& 0.93& 0.12 & 0.09 \\
        J5 & $-0.31$ & 0.41& 0.48& 0.12 & 0.09 \\
        J6 & 0.79 & 0.80 & 0.80 & 0.10 & 0.08 \\
        \hline
    \end{tabular}
\end{table}

\subsection{Power and Energy Prediction}
\label{sec:power_energy_results}

Table~\ref{tab:power_energy_accuracy} summarizes the instantaneous-power and cumulative-energy results. Power RMSE and $R^2$ use the concatenated samples of each set. The per-trial energy error $\delta_E$ is reported as mean~$\pm$~standard deviation.
\begin{table}[t]
    \caption{Power and cumulative-energy prediction accuracy.}
    \label{tab:power_energy_accuracy}
    \centering
    \footnotesize
    \renewcommand{\arraystretch}{1.12}
    \setlength{\tabcolsep}{3pt}
    \begin{tabular}{@{}p{0.47\columnwidth}ccc@{}}
        \hline
        \textbf{Evaluation set} &
        \shortstack{\textbf{RMSE}\\\textbf{(W)}} &
        \textbf{$R^2$} &
        \textbf{$\delta_E$ (\%)} \\
        \hline
        Training set ($N=66$) & 0.539 & 0.819 & $0.418\pm0.485$ \\
        Within-feature test set ($N=17$) & 0.659 & 0.683 & $0.929\pm0.700$ \\
        Unseen-feature test set P1 ($N=4$)& 0.771 & 0.432 & $0.88\pm0.75$ \\
        Unseen-feature test set P2 ($N=3$)& 0.791 & 0.738 & $1.45\pm0.96$ \\
        \hline
    \end{tabular}
\end{table}

\begin{figure}[!tp]
    \centering
    \includegraphics[width=1\columnwidth]{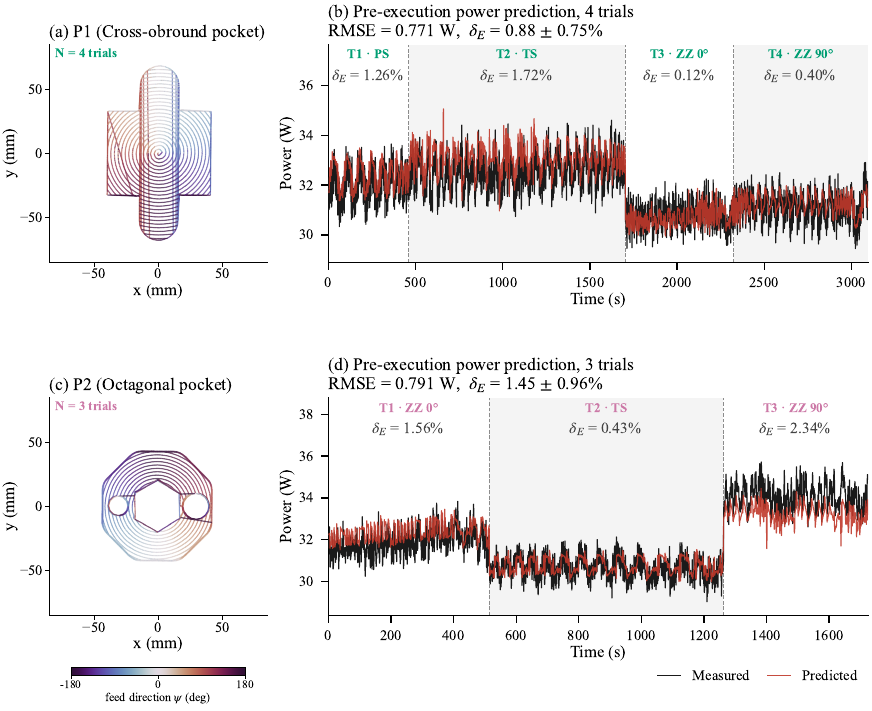}
    \caption{Pre-execution prediction on the unseen-feature test set.
(a,c) Representative P1 and P2 toolpaths colored by feed direction.
(b,d) Measured and predicted robot power. 
Power RMSE is computed over the concatenated samples of all
trials for each feature. Energy error is calculated separately
for each trial and reported as the mean $\pm$ standard
deviation across the four P1 trials or three P2 trials.}
    \label{fig:holdout_results}
\end{figure}

Figure~\ref{fig:holdout_results} shows that the predicted power follows the changes between toolpath segments. Only simulation-derived kinematics and machining parameters are used as model inputs. P1 and P2 contain unseen geometries, toolpath sequences, feed-direction patterns, and robot-posture sequences. Their similar power errors show consistent performance on two unseen features.

\subsection{Ablation and Dataset Analysis}
\label{sec:ablation_results}
\begin{figure*}[!b]
    \centering
    \includegraphics[width=0.95\textwidth]{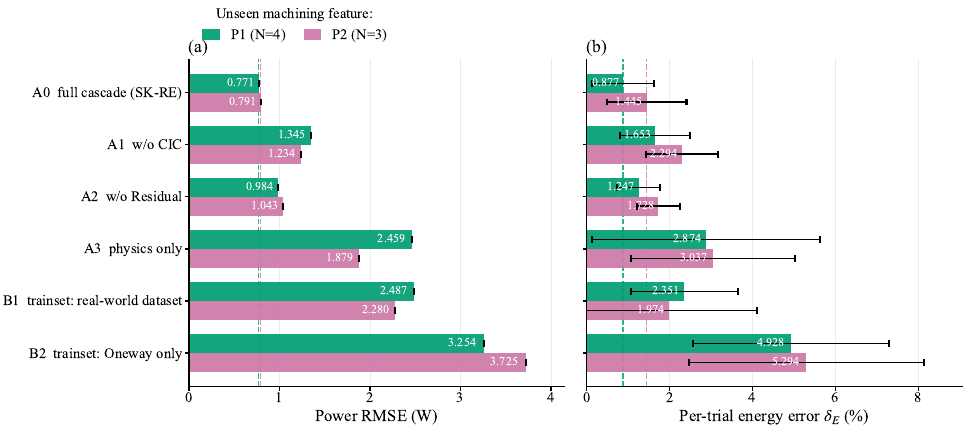}
    \caption{Ablation and training data analysis on P1 and P2. (a) Power RMSE.
    (b) Per-trial energy error. A0: full SK-RE cascade; A1: without CIC; A2:
    without residual compensation; A3: physics stage only; B1: learned stages
    trained with real-robot kinematics; B2: trained on OW only. Dashed lines
    denote the A0 results. Error bars in panel~(b) indicate the sample standard deviation of
per-trial energy errors.}
    \label{fig:ablation_results}
\end{figure*}

Figure~\ref{fig:ablation_results} compares the full SK-RE cascade (A0) with three model ablations and two training set variants. A1 removes CIC, A2 removes residual compensation, and A3 uses only the dynamics-based current estimate before power estimation. B1 trains the learned stages with real-robot kinematics but evaluates them using simulated kinematics. B2 uses only OW trials for training.

Removing either learned current stage increases power and
energy errors. A1 tests whether residual compensation can replace the explicit contact-related current estimate, while A2 tests whether dynamics and CIC alone are sufficient. Neither partial cascade matches A0, and A3 performs worse than both. These comparisons support complementary contributions from CIC and residual compensation.

B1 produces larger errors than A0, supporting the use of simulation kinematics during training when inference inputs come from simulation. B2 gives the largest errors. This comparison supports the benefit of including US trials, which add robot postures and feed directions.

\subsection{Comparison with Reference Models}
\label{sec:baseline_results}
\begin{figure}[!tbp]
    \centering
    \includegraphics[width=0.95\columnwidth]{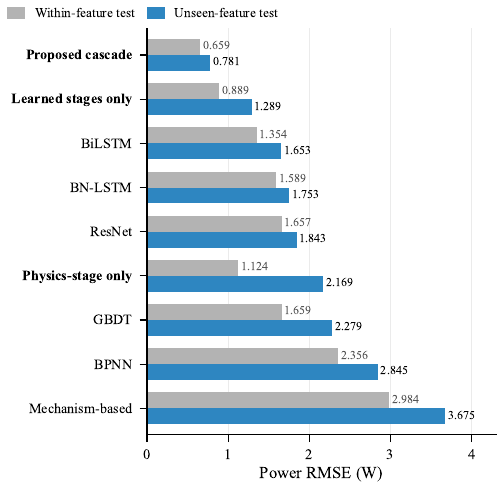}
    \caption{Power prediction accuracy of the proposed cascade and
reference models. Gray bars show RMSE of the within-feature test set.
Blue bars show the mean of power RMSE values
calculated separately for P1 and P2 in the unseen-feature test set.}
    \label{fig:baseline_results}
\end{figure}

Figure~\ref{fig:baseline_results} compares the proposed cascade with a mechanism-based reference, standalone data-driven models, and two reduced variants. The standalone data-driven references include a backpropagation neural
network (BPNN), gradient-boosted decision trees (GBDT), a batch-normalized LSTM (BN-LSTM), ResNet, and BiLSTM. All reference models use the same inputs and outputs as the proposed cascade: simulated joint kinematics and planned machining parameters as inputs, and measured robot power as the target. They are trained on the same dataset split and evaluated on the same within-feature and unseen-feature test sets. The BiLSTM reference shares the BiLSTM backbone of the cascade's learned stages but maps the inputs directly to power, without the physical current structure. The physics-stage-only variant uses the dynamics-based current followed by the power model. The learned-stages-only variant omits the dynamics-based current while retaining CIC, residual compensation, and power estimation.

Gray bars report power RMSE over all concatenated samples of the within-feature test set. Blue bars report the arithmetic mean of the RMSE values calculated separately for P1 and P2 in the unseen-feature test set. The proposed cascade gives the lowest error in both evaluations. Its advantage over the learned-stages-only variant supports the contribution of the dynamics-based current prior, while the physics-stage-only result shows the need for learned compensation.

BiLSTM is the strongest standalone data-driven reference, with corresponding errors of 1.354 and 1.653~W. The cascade reduces its unseen-feature test error by 52.8\%. Because the two models share the same backbone, this comparison supports the benefit of the current-first structure over direct power regression. The other data-driven models also show higher errors or larger increases from within-feature to unseen-feature testing. Across these comparisons, the full cascade benefits from combining a physical current prior with contact-related and residual corrections.

\section{Discussion}
\label{sec:discussion}
The current-first cascade links power prediction to the joint-current response under machining loads. The dynamics stage provides a physical current prior, while CIC combines machining parameters with the transposed Jacobian to account for the configuration-dependent transfer of cutting loads to the joints. This structure provides intermediate current estimates that can be evaluated directly. Figure~\ref{fig:current_stages} and Table~\ref{tab:current_internal_test} show that CIC provides the main correction to the dynamics-based estimate. Adding CIC
increases the $R^2$ of J1 from 0.61 to 0.91. Residual compensation
has a smaller effect on current metrics but improves
unseen-feature power and energy accuracy, as shown by its ablation in Figure~\ref{fig:ablation_results}. This is consistent with the current-squared and current-velocity features in Eq.~\eqref{eq:xpow}, through which small current corrections can affect the final power prediction. 
Together, the ablation results in Figure~\ref{fig:ablation_results} and the reference-model comparisons in Figure~\ref{fig:baseline_results} support the complementary roles of the physical prior and learned current corrections.

The contribution of SK-RE is supported by the training-input
comparison in Figure~\ref{fig:ablation_results}. B1 uses real-robot
kinematics during training and simulated kinematics during
inference, yielding power RMSE values of 2.487 and 2.280~W for
P1 and P2, respectively. These errors are approximately three
times the corresponding SK-RE results in
Table~\ref{tab:power_energy_accuracy}. This comparison supports
keeping the kinematic input source consistent between training
and pre-execution inference. SK-RE achieves this consistency while
using measured currents and power as supervisory targets.
The independently aligned machining and dry-run currents also
provide the observed CIC target. Thus, the dataset construction
allows the model to learn the physical electrical response from
inputs available during simulation-based planning.

The P1 and P2 tests provide evidence of feature-level generalization to unseen machining geometries and complete toolpath sequences. Table~\ref{tab:power_energy_accuracy} reports power RMSE values of 0.771 and 0.791~W, with per-trial energy errors of $(0.88\pm0.75)\%$ and $(1.45\pm0.96)\%$, respectively. Figure~\ref{fig:holdout_results} further shows that the predicted power follows the changes between toolpath segments.
These results were obtained on a UR10 using a 3~mm two-flute
flat-end mill for low-load wax machining. Beyond this setting,
the $n$-DOF dynamics formulation in
Eqs.~\eqref{eq:rbd}--\eqref{eq:idyn} and the load-to-current
relation in Eq.~\eqref{eq:cic} provide a basis for adapting the
cascade to other robot arms, tools, and workpiece materials.
Such adaptation can retain the current-first structure and SK-RE
procedure, with updated robot dynamics, drive gains, and learned
stages trained on representative machining data.

\section{Conclusion}
\label{sec:conclusion}
This study presented a Jacobian-aware, current-first cascade for pre-execution robot energy prediction in milling. The framework combines rigid-body dynamics, CIC prediction, residual compensation, and a linear power model. SK-RE pairs simulated kinematics with
measured electrical targets, allowing inference from planning-level inputs. On P1 and P2 features, power RMSE reached 0.771 and 0.791~W, with mean per-trial energy errors of 0.88\% and 1.45\%.
Ablations support the contributions of the learned current stages and simulation-derived training inputs. The framework provides an upstream energy estimate for evaluating planned machining trajectories. Future work will extend the evaluation to other robots and machining conditions.

\appendices
\section{Preliminary Experiment on Configuration-Dependent Robot Energy}
\label{app:preliminary}

\begin{figure}[!tbp]
    \centering
    \includegraphics[width=0.95\columnwidth]{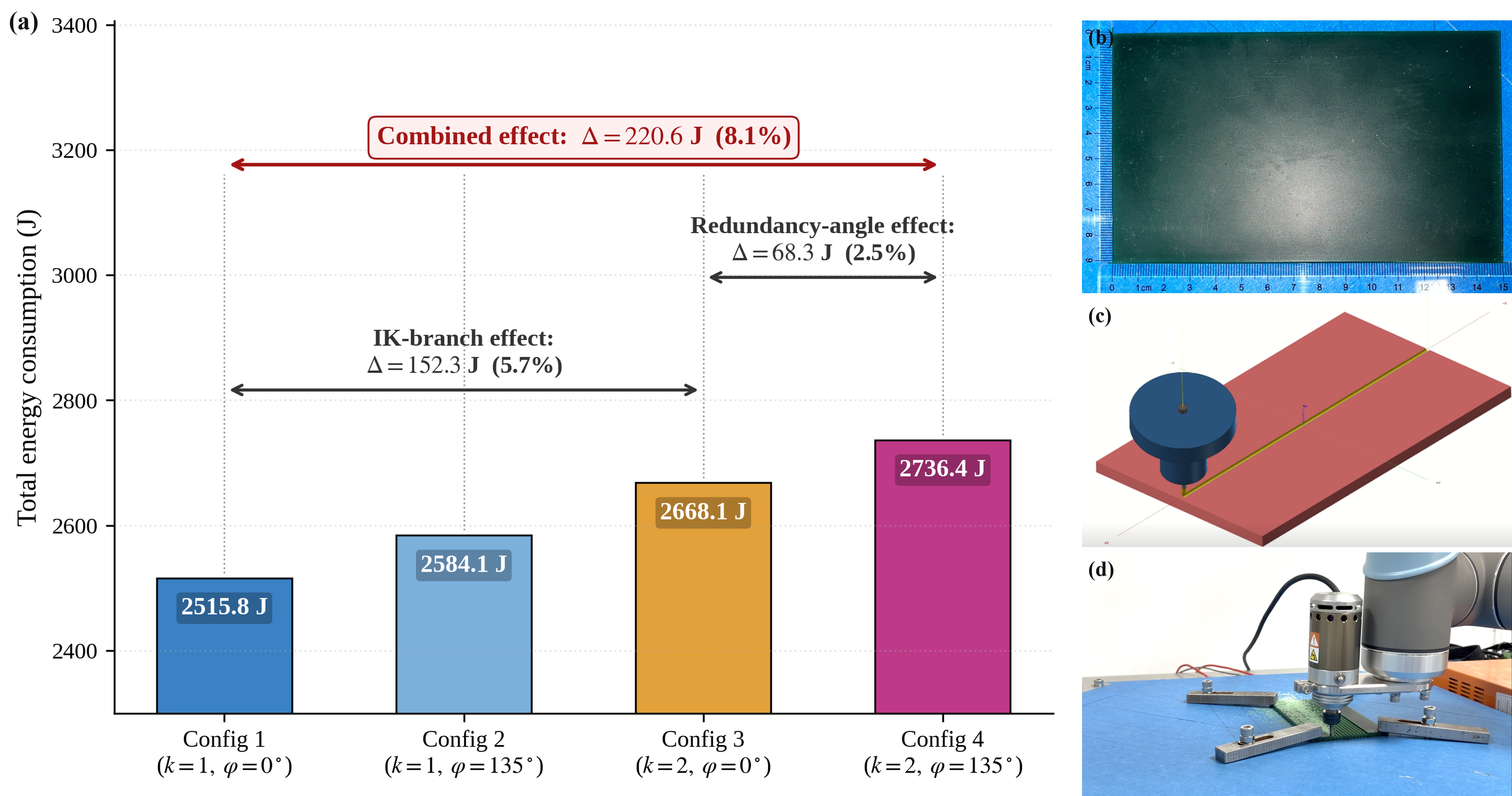}
    \caption{Preliminary configuration comparison for UR10 slot milling.
    (a) Measured robot energy for four trajectories with the same
    TCP path and tool-axis direction but different IK branches $k$
    and tool-axis rotation angles $\varphi$.
    (b) Machining-wax workpiece.
    (c) Simulated 150~mm slot-milling toolpath.
    (d) Physical milling setup.}
    \label{fig:redundancy}
\end{figure}

Four trajectories were tested for a 150~mm planar slot-milling task
on the UR10 using a 3~mm flat-end mill and machining wax. All four trajectories used the same TCP path, tool-axis direction, spindle speed, feed rate, and depth of cut. They differed in the
initial inverse-kinematics (IK) branch $k$ and the rotation angle $\varphi$ about the tool axis. The four configurations were
$(k,\varphi)=(1,0^\circ)$, $(1,135^\circ)$, $(2,0^\circ)$, and
$(2,135^\circ)$. These configurations performed the same prescribed machining task through different joint-space trajectories.
Figure~\ref{fig:redundancy} summarizes the energy comparison, workpiece, simulated toolpath, and physical setup. Robot electrical energy was obtained by integrating the measured robot power for each trajectory.

The measured energies were 2515.8, 2584.1, 2668.1, and 2736.4~J, respectively. At $\varphi=0^\circ$, changing the IK branch from
$k=1$ to $k=2$ increased energy by 152.3~J. At $k=2$, changing
$\varphi$ from $0^\circ$ to $135^\circ$ increased energy by 68.3~J.
The difference between the highest- and lowest-energy configurations was 220.6~J. Selecting the lowest-energy configuration therefore corresponds to an energy reduction of approximately 8.1\% relative
to the highest-energy case:
\begin{equation}
    \frac{2736.4-2515.8}{2736.4}\times100\%
    \approx 8.1\%.
    \label{eq:preliminary_energy_reduction}
\end{equation}

These observations show that the prescribed TCP path and cutting parameters alone do not determine robot energy consumption. Different joint-space configurations change the robot loading and
the transmission of cutting loads through the Jacobian.
The experiment therefore motivates pre-execution energy prediction for comparing joint-space trajectories that perform the same machining task.

\section*{Acknowledgment}
The authors thank the participating institutions for providing the research facilities and technical support for the experimental work.

\section*{Declarations}

\begin{itemize}
\item \textbf{Conflict of interest}: The authors declare no conflicts
  of interest.
\item \textbf{Data availability}: The experimental datasets supporting
  the findings of this study will be made available upon reasonable
  request.
\item \textbf{Code availability}: The implementation code will be made
  available at a public repository upon acceptance.
\end{itemize}

\end{document}